\documentclass{article}

\usepackage{microtype}
\usepackage{graphicx}
\usepackage{subfigure}
\usepackage{booktabs} 

\usepackage{hyperref}

\usepackage[accepted]{icml2024}

\usepackage{amsmath}
\usepackage{amssymb}
\usepackage{mathtools}
\usepackage{amsthm}

\usepackage{amsmath,amsfonts,bm}

\def\eqref#1{equation~\ref{#1}}

\def\1{\bm{1}}

\DeclareMathAlphabet{\mathsfit}{\encodingdefault}{\sfdefault}{m}{sl}
\SetMathAlphabet{\mathsfit}{bold}{\encodingdefault}{\sfdefault}{bx}{n}

\def\sV{{\mathbb{V}}}

\usepackage[capitalize,noabbrev]{cleveref}

\theoremstyle{plain}

\theoremstyle{definition}

\theoremstyle{remark}

\usepackage{booktabs}
\usepackage[export]{adjustbox}
\usepackage{wrapfig}
\usepackage{multicol}
\usepackage{subcaption}
\usepackage{multirow}
\usepackage{enumitem}
\usepackage{graphicx,graphbox}
\usepackage{xspace}
\definecolor{myorange}{HTML}{FF9933}
\definecolor{mygreen}{HTML}{85A177}
\newcommand{\token}[1]{$\langle$\texttt{#1}$\rangle$}
\newcommand{\Algo}{\textsc{Tag-LLM}\xspace}

\usepackage[textsize=tiny]{todonotes}

\icmltitlerunning{Tag-LLM: Repurposing General-Purpose LLMs for Specialized Domains}

\begin{document}

\twocolumn[
\icmltitle{Tag-LLM: Repurposing General-Purpose LLMs for Specialized Domains}




\begin{icmlauthorlist}
\icmlauthor{Junhong Shen$^*$}{yyy}
\icmlauthor{Neil Tenenholtz}{comp}
\icmlauthor{James Brian Hall}{comp}
\icmlauthor{David Alvarez-Melis}{comp,sch}
\icmlauthor{Nicol\`{o} Fusi}{comp}
\end{icmlauthorlist}

\icmlaffiliation{yyy}{Carnegie Mellon University}
\icmlaffiliation{comp}{Microsoft Research}
\icmlaffiliation{sch}{Harvard University}

\icmlcorrespondingauthor{Junhong Shen}{junhongs@andrew.cmu.edu}

\icmlkeywords{Machine Learning, ICML}

\vskip 0.3in
]



\printAffiliationsAndNotice{\textsuperscript{*}Work done during internship at Microsoft Research.} 

\begin{abstract}
Large Language Models (LLMs) have demonstrated remarkable proficiency in understanding and generating natural language.
However, their capabilities wane in highly specialized domains underrepresented in the pretraining corpus, such as physical and biomedical sciences. This work explores how to repurpose general LLMs into effective task solvers for specialized domains.
We introduce a novel, model-agnostic framework for learning custom \textit{input tags}, which are parameterized as continuous vectors appended to the LLM's embedding layer, to condition the LLM.
We design two types of input tags: \textit{domain tags} are used to delimit specialized representations (e.g., chemical formulas) and provide domain-relevant context; \textit{function tags} are used to represent specific functions (e.g., predicting molecular properties) and compress function-solving instructions. We develop a three-stage protocol to learn these tags using auxiliary data and domain knowledge.
By explicitly disentangling task domains from task functions, our method enables zero-shot generalization to unseen problems through diverse combinations of the input tags. It also 
boosts LLM's performance in various specialized domains, such as predicting protein or chemical properties and modeling drug-target interactions,
outperforming expert models tailored to these tasks.
\end{abstract}
\section{Introduction}
\label{sec:intro}
\looseness=-1
Large Language Models (LLMs) excel at processing and generating text across diverse topics due to their general-purpose design. However, their performance degrades in specialized domains underrepresented
in the pretraining corpus. These domains range from natural language with non-standard lexical distributions, such as health records, to non-linguistic, symbolic representations, such as DNA and protein sequences. For instance, general-purpose LLMs like LLaMA \citep{llama} and GPT-4 \citep{GPT4}  have demonstrated notable deficiencies in processing amino acid sequences for proteins (e.g., \texttt{MTVPDRSEIAG}) \citep{evidence1} and SMILES strings for chemical compounds (e.g., \texttt{CCO[C@H](C(=O)O)}) \citep{guo2023indeed}, hampering their adoption for a wide range of scientific problems.

\looseness=-1
To address this issue, various specialized LMs have been developed in fields such as disease diagnosis \citep{evidence2}, chemical synthesis \citep{Chilingaryan2022BARTSmilesGM}, and drug discovery \citep{Vinod2023ReprogrammingPL}. While these models perform well, they are trained from scratch, requiring a significant amount of compute and in-domain data. To reduce the development cost and take advantage of the capacity of general LLMs, recent work has proposed to fine-tune existing LLMs or perform in-context learning, but the former is prone to catastrophic forgetting and can hurt the model's reasoning abilities \citep{clinical, quantitative}, while the latter is less effective and robust \citep{evidence3, lu2022fantastically}. We thus ask: \textit{can we effectively repurpose general-purpose LLMs for specialized tasks without compromising their linguistic and reasoning capabilities?}

\begin{figure*}[t]
  \centering
 \includegraphics[width=0.95\textwidth]
 {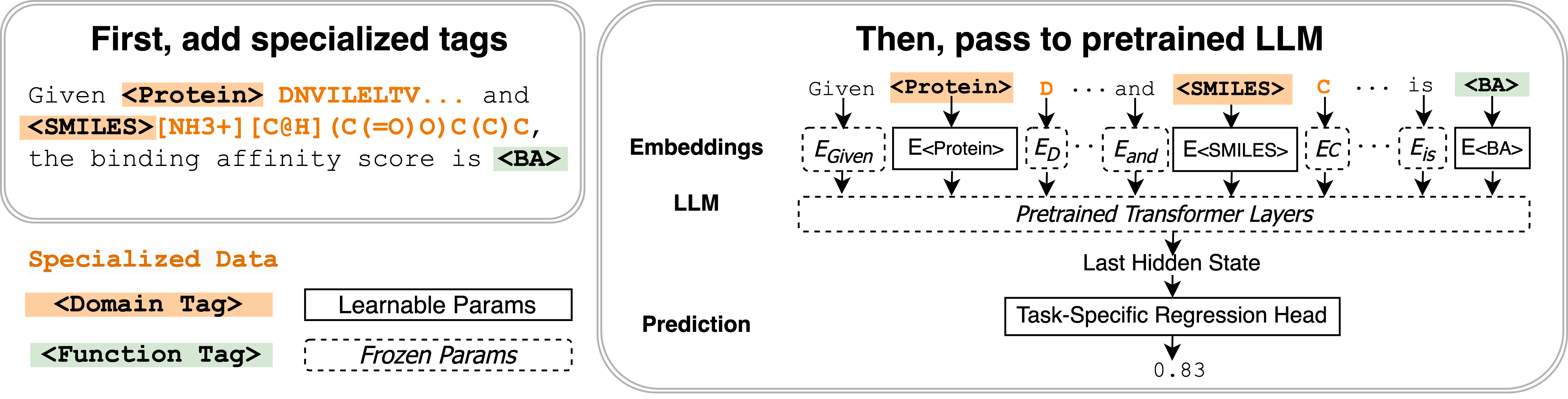}
  \caption{\small Using the task of protein-drug binding affinity prediction as an example, our method injects \textcolor{myorange}{domain tags} \token{Protein}, \token{SMILES} and a \textcolor{mygreen}{function tag} \token{Binding Affinity} to the input, which are mapped to specially trained embeddings. The model's last hidden state is passed to a task-specific head to generate  predictions of the desired type (e.g., a \textit{scalar} binding affinity value in this case).}
\label{fig:example}
\end{figure*}

Our proposed solution is \Algo,
a modular system of meta-linguistic input tags that condition the LLM for domain- and task-specific behavior 
(Figure~\ref{fig:example}). These input tags are parameterized as continuous vectors 
\textit{appended} to the LLM's embedding layer, preserving its original weights and language capabilities. 
The idea of adapting models via learnable embeddings has been partially explored by prompt tuning~\citep{prompttuning} and prefix tuning methods \citep{prefixtuning}, which learn soft prompts prepended to the input. However, these methods require learning a dedicated set of parameters for \textit{every single task}, resulting in higher computational costs and limited reusability. The prompts are trained directly on the target task, and do not distinguish between task-specific and general domain knowledge. Once learned, they affect the entire input globally.
In contrast, our approach seeks to learn modifiers that are reusable across tasks, have fine-grained localized effects on the input, and can be trained using more general types of in-domain data.

Our method enables generalization to new settings by disentangling \textit{domains} (e.g., `French language' or `protein sequence') from \textit{tasks} (e.g., `translation' or `binding affinity prediction'), using distinct procedures to learn and encode each. Specifically, we design two types of tags: \textit{domain tags} are used to signal the start of specialized data and specify their domains;
\textit{function tags}, on the other hand, pertain to the abilities that can be shared across domains and instruct the LLM on how to solve a family of relevant tasks.
Hence, different combinations of domain and function tags can be used to represent various domain-grounded tasks. Once we have a collection of learned tags, we can use them to solve new problems, much like assembling pieces of a puzzle.

By design, domain and function tags apply to subsets of the input. Moreover, as they are not tied to a single task, we can take advantage of auxiliary datasets and perform multi-task training to improve their efficacy. Specifically, we introduce a three-stage training protocol for these speicialized tags:
\begin{itemize}[leftmargin=*,itemsep=1pt,parsep=0pt,topsep=0pt,partopsep=0pt]
    \item \textbf{Stage 1}:\textbf{Domain tags} are trained for next-token prediction using auxiliary, unsupervised, in-domain data. These tags encode general domain characteristics learned from token distributions and can be used for `priming' the LLM. 
    \item \textbf{Stage 2}: \textbf{Single-domain function tags} are trained using  supervised single-domain data, with the domain tags inserted into the input.
    We allow the domain tags to be updated to enrich them with task-relevant knowledge.
    \item \textbf{Stage 3}: \textbf{Cross-domain function tags} are trained 
    using  supervised cross-domain data. We formulate a multi-task setting where each function tag is combined with multiple domain tags to represent a family of relevant tasks. The goal is to
    extract shared abilities from these tasks.
\end{itemize}
This three-stage procedure naturally leverages specialized domain data with varying levels of availability and supervision (e.g., unlabeled raw data vs. task-specific input-output pairs), as we show in Figure~\ref{fig:pyramid}. Also, it leads to better usage of domain knowledge and improves performance over naive prompt tuning. The modularity of the tagging system allows us to add specialized tags incrementally, gradually expanding the LLM's capabilities to new tasks and domains.

We also observe that many tasks in specialized domains involve predicting non-textual outputs (e.g., scalar values for regression or probability vectors for classification). In response, we augment the function tags with function-specific linear heads. This allows us, for example, to directly predict a scalar-valued affinity score for drug discovery instead of having the model generate decimals one digit at a time, which LLMs often struggle with \citep{hendrycks2021measuring}.

We evaluate \Algo on a diverse set of ten domains, encompassing eight languages and two specialized scientific domains (protein sequences and SMILES molecule representations). We chose the latter two because they represent non-linguistic, highly specialized domains with distributions significantly different from natural language, and they are relatively well understood with established baselines. We experiment with the LLaMA-7B model \citep{llama}. Our results demonstrate that the domain tags can act as effective context switchers,
and a  function tag can be applied to multiple domains to solve different tasks, achieving zero-shot generalization to unseen problems.
We show that \Algo improves LLaMA's performance on ten translation tasks, matching specialized multilingual LLMs. We also achieve state-of-the-art results on multiple drug discovery datasets from the Therapeutics Data Commons benchmark \citep{TDC}.
Given its model-agnostic and plug-and-play nature, \Algo  provides a systematic way to leverage the rapid advancements in LLMs to tackle a wider range of real-world problems. Code is available at \href{https://github.com/sjunhongshen/Tag-LLM}{https://github.com/sjunhongshen/Tag-LLM}.

\section{Related Work}
\label{sec:related_work}

\subsection{LLM Adaptation}
\label{sec:related:adaptation}

Our work aims to adapt general-purpose LLMs to specialized tasks, including non-linguistic ones. Although a similar goal motivated previous work for NLP tasks, we differ from them in three aspects, which we discuss in detail below.

\textbf{Domain \& Task Type } Existing domain adaptation work \citep{CTRL, generation1, generation2, generationsurvey} mainly considers NLP domains defined in terms of content topics (e.g., business, sports, music), literary forms (e.g., poetry, story), or linguistic styles (e.g., polarity, formality). However, we expand beyond these NLP domains into non-linguistic ones, such as  molecule formulas and protein sequences. 
We also generalize to more complex problems where the target outputs consist of a mixture of non-text elements, such as scalar-valued scores.

\looseness=-1
\textbf{Adaptation Strategy } Due to the semantic gap between natural language and specialized non-NLP representations, existing adaptation methods are insufficient. In particular, full fine-tuning is computationally costly and requires substantial labeled data.
Hard-prompting with instructions or demonstrations \citep{Brown2020Language} does not perform gradient updates.
However, it often requires cumbersome trial-and-error prompt crafting and is highly sensitive to format changes in the prompt \citep{lu2022fantastically}. Meanwhile, the amount of  information included in the prompt is limited by the LLM's context length. 
Different from them, we learn special tags that compress large amounts of domain or task knowledge from auxiliary data into a few embedding parameters, reducing the cost of fine-tuning or manual prompting.

\looseness=-1
\textbf{Modular Design } Our method has a similar motivation to parameter-efficient fine-tuning (PEFT) techniques such as LoRA \citep{lora}, prompt tuning \citep{prompttuning}, and prefix tuning \citep{prefixtuning}. These methods offer a middle ground between full fine-tuning and hard-prompting, modifying fewer parameters while being less limited by context length. Nonetheless, existing approaches can typically adapt to a single downstream task at a time and learn distinct sets of parameters even for tasks that share knowledge. In contrast, our method separates task domains from task functions---we use modularized tags to represent different domains/functions and learn them via a hierarchical training protocol. This leads to better reusability and sample efficiency. We also show empirically that \Algo significantly outperforms prompt tuning (see Section~\ref{sec:exp:science}) , which  provides further evidence that our method differs from PEFT methods not only conceptually, but also functionally.

\subsection{Conditional Generation \& Task Compression}
Our input tags are trained to condition the LLM. Prior work has explored training models with conditioners from scratch \citep{CTRL} or using pretrained models as we do. In the latter case, methods based on hard \citep{hardpromptmarker} or soft prompts~\citep{softpromptmarker,prefixmarker2, prefixmarker} have been developed. These conditioners are ``global" and apply to the entire input, whereas our specialized tags can be inserted into different places of the input for different purposes. Additionally, our method leverages a hierarchical structure  with more specialized tags building upon more general ones (see Figure~\ref{fig:pyramid}), unlike most existing conditioning methods that are non-hierarchical.

Function tags can also be viewed as a form of task compression. Several recent works have explored task or instruction abstraction. For instance, \citet{gist} studies how the LLM itself can summarize a user-supplied instruction into a soft prompt for computational efficiency. Unlike that work,  we do not require explicit text instructions but learn the function tags entirely from the data. \citet{composition} represents each NLP task in the T0 benchmark \citep{T0}
as a combination of discrete latent codes and examine the shared codes across tasks. Instead of performing such post-hoc analyses, we separate the required ability from the input properties (i.e., domains) during training. This gives us better control in terms of task decomposition.  

\subsection{LLMs for Specialized Non-Linguistic Domains}
Extensive efforts have been devoted to developing domain-specific models for specialized tasks such as coding \citep{code2}, mathematical reasoning \citep{math}, protein structure prediction \citep{alphafold}, and gene effect prediction \citep{nasbench360, shen2022dash}. Yet these models typically have custom architectures and are trained from scratch, requiring large amounts of in-domain data and compute. Here we seek a method that can leverage existing LLMs and requires minimal in-domain data, allowing for faster on-demand adaptation by non-domain-experts.
In a similar vein, several recent works adapt pretrained LLMs to non-NLP fields such as physical and biological sciences \citep{ clinical, Vinod2023ReprogrammingPL, orca, shen2024ups}, tabular data \citep{LIFT, Roberts2021AutoMLDD}, and quantitative reasoning \citep{quantitative}. These methods require either fine-tuning the entire model \citep{FPT,orca} or  manually converting non-text data into text \citep{LIFT}. In contrast,  we learn continuous embeddings that condition the model with domain-specific information, which is more parameter-efficient than fine-tuning and  more effective than text conversion (see experimental results in Section~\ref{sec:exp:science}).

\section{Methods}
\label{sec:methods}
In this section, 
we introduce two types of input tags (Section~\ref{sec:design}) and provide the intuition behind their design. Then, we discuss the three-stage training protocol (Section~\ref{sec:training}).

\subsection{Input Tags: Design and Parametrization}
\label{sec:design}

We consider the problem of repurposing a general-purpose LLM to solve tasks 
in specialized domains, i.e., domains that are
underrepresented in the model's pretraining corpus but for which we have external data available for adaptation. We are especially interested in technical and rapidly evolving domains, such as those at the frontiers of science, where new, previously unseen data is constantly being generated, and knowledge is represented through a mixture of linguistic and non-linguistic data. These two characteristics (namely, unobserved knowledge 
and heterogeneous data formats) make pretrained LLMs ill-suited for solving specialized problems without further modification.

\looseness=-1
As discussed in Section~\ref{sec:related:adaptation}, existing adaptation techniques have limitations: full fine-tuning is computationally costly, while PEFT methods are task-specific and non-transferable to new problems. 
To address these challenges, we seek to enhance the model's vocabulary with small sets of parameters pertinent to both individual tasks and broader domains. We refer to these parameters as ``input tags."

To distinguish between general domain information and task-specific instructions, we introduce two types of input tags: \emph{domain tags} and \emph{function tags}. The domain tags delimit the occurrence of domain-specific data (e.g., French or protein sequences) in the input and encode domain-level information that can be relevant to multiple downstream tasks. Function tags, in turn, guide the model to use these domain inputs to solve a family of tasks (e.g., translation or binding affinity prediction) by encoding task semantics.

Both tags are parametrized in the same way. Given an LLM with pretrained vocabulary $\sV$ and embedding dimension $d$, we instantiate an input tag as a learnable parameter in $\mathbb{R}^{p\times d}$, where $p$ denotes the number of tokens used for each tag and is a tunable hyperparameter. Note that the tags exist in the embedding space, and we do not assign actual words to them, so the model cannot output a tag. We initialize the tag using the model's average embedding $\hat{v} = \frac{1}{|\sV|}\sum_{v\in \sV} v$. More specifically, we first scale $\hat{v}$ by $\frac{1}{|\sV| \|\hat{v}\|}\sum_{v \in \sV}\|v\|$  to match its norm with the average norm of the ordinary text token embeddings in $V$. 
Then, we stack $p$ copies of the re-scaled $\hat{v}$ to initialize a tag.

\subsection{Training Protocol}
\label{sec:training}
\begin{figure}
  \centering
\includegraphics[width=0.46\textwidth]{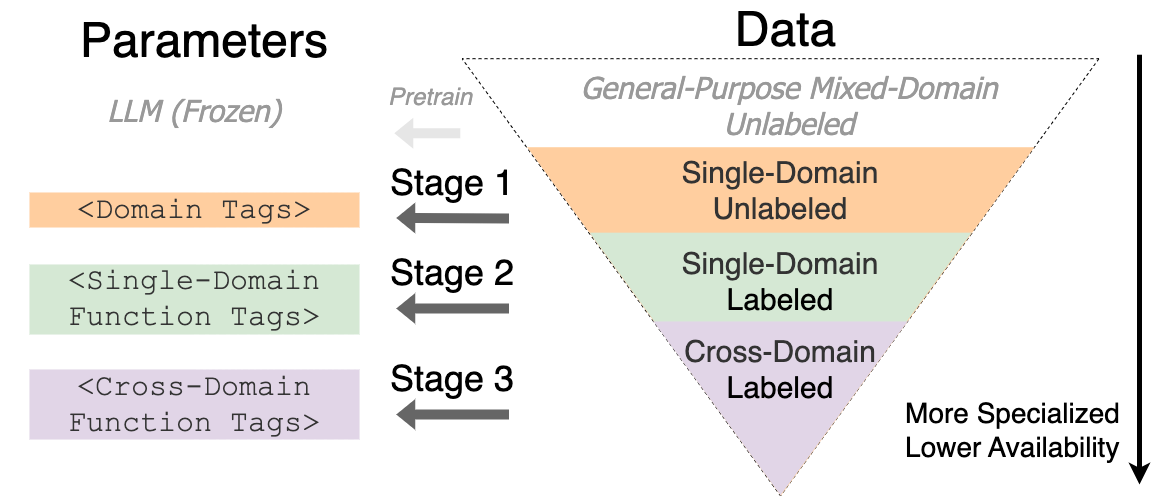}   
  \vspace{-2mm}
  \caption{\small Specialized data are not all created equal. We exploit different levels of data availability in our 3-stage training protocol.}
  \vspace{-4mm}
 \label{fig:pyramid}
\end{figure}

Since domain and function tags have distinct natures and roles, we train them using different types of data. Domain tags, which are more general and task-agnostic, can be trained using unlabeled data from a single specialized domain (e.g., amino acid sequences for representing proteins). Such unsupervised data is often available in relatively large amounts. Function tags, on the other hand, require labeled data specific to the set of tasks they encode (e.g., property prediction based on a protein sequence), which is scarcer. Function tags that involve multiple domains (e.g., predicting the binding affinity between a protein and a drug) require multi-domain labeled data, the scarcest kind. 
Therefore, we design a three-stage hierarchical protocol that develops input tags progressively, from general to specialized (Figure~\ref{fig:pyramid}).

\subsubsection*{Stage 1: single-domain domain tags}\label{sec:training:stage1}
Our intuition is that tags for different domains 
should summarize domain-relevant  knowledge. Then, inserting them in the input should provide the context more efficiently---we can shift the LLM's conditional distribution  at the moment a domain tag appears, rather than having the model determine the context itself after processing the entire input.
To this end, we insert the domain tags in front of specialized data.  Denote the domain tag as $M\in \mathbb{R}^{p\times d} $. Given tokens in specialized domains $\{x_1, \cdots ,x_n\}$ whose embedding matrix is $X_e \in\mathbb{R}^{n\times d}$, we prepend $M$ to $X_e$ so that the embedded input used to query the LLM is $[M; X_e] \in\mathbb{R}^{(p+n)\times d}$. 

\textbf{Training } We train the domain tags using in-domain demonstrations and self-supervised next-token prediction. That is, $M$ is learned to optimize
$l_{M} := \sum_{t\in[n]}\mathbb{P}(x_{t+1}|[M;x_{1:t}])$. 
We use the standard autoregressive mask for training and inference.
At later stages, domain tags can  be further fine-tuned  (``enriched") along with the function tags to encapsulate more task-relevant knowledge. To ensure that domain tags maintain their role during enrichment, we also compute and optimize $l_{M}$ from the specialized data.

\textbf{Effect } The benefit of stage 1 is twofold. First, it allows us to inject general domain knowledge into the prompting process to improve downstream performance, which we show via ablation studies in Section~\ref{sec:ablation_ability}. Second, this stage contributes to the generalization ability of \Algo, since domain tags are task-agnostic.

\subsubsection*{Stage 2 \& 3: single/cross-domain function tags} \label{sec:training:stage2_3}

The goal of function tags is to encode task-specific abilities. However, rather than simply summarizing user-provided instructions as is done in prior work \citep{gist}, our function tags learn the essence of the tasks entirely from the data. To achieve this, we place them at the end of the input so that they can attend to both the specialized data and the domain tags. Formally, given a labeled dataset $\{(X, y)\}$, we first preprocess the data by inserting the domain tags $M$ into \(X\) to prefix specialized data. Then, we append the function tag $F$ to the end. Prediction is based on the last hidden state of the LLM (corresponding to $F$), and can take several forms (see Section~\ref{sec:beyond_text}).

\textbf{Training } 
The function tags are learned by directly optimizing the downstream task loss $l_{F}$. For instance, for text generation, \(l_{F}\) is the standard cross-entropy loss; for regression-based tasks, \(l_{F}\) can be the mean-squared-error between the model output and the label.
We also differentiate between single-domain functions (learned in stage 2) and cross-domain functions (learned in stage 3). The former applies to one specific domain and is trained on a single dataset, so the domain tag is allowed to be enriched. For cross-domain functions, there are either multiple domains appearing in the same input (e.g., $F(A, B)$ where $A$ and $B$ are different domains), or each task involves one domain, but the same function is required by multiple tasks (e.g., $F(A)$ and $F(B)$). In both cases, the domain tags are frozen to avoid being polluted by out-of-domain information. When a function is applicable to multiple domains (i.e., $F(A)$ and $F(B)$), we formulate a multi-task setting by concatenating multiple datasets and train $F$ to optimize $l_{F}$ on all datasets.

\textbf{Effect } Stage 2 \& 3 allow the model to learn task knowledge from feature-label pairs (without human instructions) and compress them into a few embedding parameters. This boosts the empirical performance (see ablation studies in Section~\ref{sec:ablation_ability}). In Section~\ref{sec:exp:trans}, we will also show how a cross-domain function tag can be combined with various language domain tags and generalize to new tasks, using multilingual translation as an example. Hence, learning the function tags contributes to our method's zero-shot generalization ability.

In general, \Algo not only enables more granular control over the LLM  but also allows composing multiple tags for solving different and potentially unseen problems. Meanwhile, new tags can easily be added to the system, regardless of whether they depend on existing tags or not. 

\subsection{Beyond Text Generation } 
\label{sec:beyond_text}

The input tags defined so far adapt the LLM at the \textit{input level}. However, specialized domains (especially scientific ones) often involve tasks whose output is not naturally represented as text, such as scalar-valued scores or vector-valued probability distributions. The naive solution is to have the model generate text-based outputs and  convert to the desired format afterward. However, this approach is fraught with problems.  First, the cross-entropy (CE) loss is not a good proxy for regression losses like mean-squared-error (MSE),
because the former is agnostic to digit order. For instance, $\textup{CE}(``\texttt{3.14}", ``\texttt{3.24}")= \textup{CE}(``\texttt{3.14}", ``\texttt{3.15}")$ since both pairs of numbers differ by one digit, but $\textup{MSE}(3.14, 3.24) \neq \textup{MSE}(3.14, 3.15)$.
Moreover, certain constraints on the output, such as $\|h\|_2=1$, cannot be easily enforced when the output is in text form.

To overcome these limitations and extend the capacity of LLMs beyond text generation, we pair the function tags related to numerical prediction with task-specific regression heads. Formally, given a task $t$, we initialize $w_t \in R^{d\times d_t}$, where $d_t$ is the desired output dimension (e.g., $1$ for univariate regression, $k$ for $k$-class classification). The prediction is obtained by multiplying $w_t$ with the model's last hidden state, and can be scored using any desired loss such as MSE. We show in Section~\ref{sec:ablation_regression} that task-specific output modeling can significantly improve the downstream performance.

\section{Specialized NLP: Multilingual Translation}
\label{sec:exp:trans}

\looseness=-1
In the remainder of this paper, we provide empirical evidence that \Algo can effectively repurpose LLMs for specialized domains. We will start with a multilingual translation task and dive into non-linguistic domains later. 
In all experiments, we leverage the pretrained LLaMA-7B model \citep{llama} with embedding dimension $d=4096$. We set the tag length $p=10$, resulting in  $40960$ trainable parameters for each input tag. We run our experiments on a single NVIDIA A100 GPU. 
More experiment details can be found in Appendix~\ref{appendix:ours}.

For multilingual translation, the domain tags are used to represent different languages, and we train a shared function tag \token{Translate} to encode the translation ability. Our goal is to verify (1) the domain tags can effectively extract domain information from the data; and (2) the function tag can generalize to unseen domains and translation pairs.

Using the OPUS-100 datasets \citep{opus}, we first learn 8 different language tags, namely \token{EN}, \token{FR}, \token{RU}, \token{DE}, \token{IT}, \token{EL}, \token{ES}, and \token{PT}, each trained separately using next-token prediction. Then, we take 5 paired datasets, namely \textsc{en}-\textsc{fr},\textsc{en}-\textsc{ru}, \textsc{en}-\textsc{de}, \textsc{en}-\textsc{it}, and \textsc{en}-\textsc{el}, combine them, and format each example as follows:

{\footnotesize
\vspace{-0.7cm}
\begin{align*}
  &  \texttt{Input: \!\!\!\textcolor{myorange}{\token{source\_lang}} source\_text} \\ 
 &  \texttt{Output: \!\!\!\!\textcolor{myorange}{\token{target\_lang}} \textcolor{mygreen}{\token{Translate}} target\_text}
\end{align*}}%
We train \token{Translate} on the combined dataset using the cross-entropy loss over \texttt{target\_text} (multi-tasking). Note that the training data involves only 6  of the 8 languages (excluding \textsc{es} and \textsc{pt}). That is, there are 6 domains and 5 language pairs ``seen'' by \token{Translate}.

\begin{table*}
\Large
\caption{\small spBLEU scores on Flores-101 datasets \citep{flores}, with baseline values taken from \citet{xglm,bloom}.  In general, our method achieves performance comparable to that of the  multilingual specialized baselines, and the input tags successfully generalize to previously unseen combinations, which proves the compositionality of our framework. }
\vspace{-1mm}
\label{table:translate}
\centering
\renewcommand{\arraystretch}{1.1}
\resizebox{\textwidth}{!}{\begin{tabular}{lcccccccccc}
		\toprule
  &\multicolumn{6}{c}{Seen Language Pairs}&\multicolumn{2}{c}{Seen Domains / Unseen Pair}& \multicolumn{2}{c}{Unseen Domains} \\
   \cmidrule(lr){2-7} \cmidrule(lr){8-9} \cmidrule(lr){10-11} 
   & \textsc{en-fr} & \textsc{fr-en}&\textsc{en-ru} &\textsc{ru-en}&\textsc{en-de}&\textsc{de-en}&\textsc{it-fr}&\textsc{fr-it}&\textsc{es-pt}&\textsc{pt-es}\\  
\midrule
\textbf{Specialized}\\
{M2M}&42.0& 37.2&\textbf{27.1}&27.5&\textbf{32.6}&35.8&\textbf{34.4}&\textbf{28.6}&28.1&26.9\\
 {XGLM}&36.0&40.4&24.2&\textbf{30.4}&27.6&38.8&-&-&-&-\\
{BLOOM}&\textbf{45.0}&45.6&-&-&-&-&31.4&24.0&\textbf{29.1}&\textbf{28.1}\\
		\midrule  
	 \textbf{General-Purpose}\\
{GPT-3}&36.1&42.8 & 11.2&28.1&25.9&40.4&-&-&-&-\\
  {Tag-LLaMA \textbf{(Ours)}} &40.5&\textbf{45.7}&14.2&30.1&23.0&\textbf{40.5}&28.8&23.9&24.6&27.8\\
\bottomrule
	\end{tabular}}
 \end{table*}
 
Similar to \citet{xglm,bloom}, we perform evaluation on a subset of the Flores-101 devtest set \citep{flores}. We compute the sentence-level SentencePiece BLEU and compare with the following LLM baselines: the translation-task-specific M2M \citep{m2m}, the more general multilingual pretrained XGLM \citep{xglm} and BLOOM \citep{bloom}, and the most general GPT-3 \citep{Brown2020Language}.  Note that our  goal is not to beat the state-of-the-art, since LLaMA is mainly trained on English.
Instead, we are interested in studying whether the learned tags can  encode the  translation task \textit{from scratch} without extra context (note that our prompt  consists only of the tags and the sentence pairs, but not  instructions). 

The results are shown in Table~\ref{table:translate}. Notably, our method matches the multilingual baselines on seen domains, ranking first in two settings. This shows that the input tags can effectively compress domain- and task-relevant information.
We also observe that all general-purpose models (ours and GPT-3) are better at generating English than other languages, likely due to its predominance in the pretraining corpus. In addition, our method yields reasonable performance for the task \textsc{es}$\leftrightarrow$\textsc{pt} (last columns in Table~\ref{table:translate}), even though  \token{Translate} has never  seen either of the languages before. This shows that the function tags can  extract shared property from multiple tasks and generalize to unseen tasks. This allows us to reuse learned tags for new problems, which previous prompt tuning methods are not capable of.

Despite the strong performance, the translation task does not fully showcase the potential of our method, since it requires only language reasoning, which is an inherent ability of LLMs. In the next section, we tackle  more challenging tasks outside of the NLP domains, which require specialized knowledge not present by default in general-purpose LLMs.

\begin{table*}
\Large
\centering
  \caption{\small Performance of 
  \Algo vs. baselines on   descriptor prediction,  QED prediction, and two TDC datasets \citep{TDC} ($\downarrow$$\setminus$$\uparrow$: lower$\setminus$higher is better). 
  \Algo significantly outperforms the PEFT baselines, showing the effect of adapting LLMs to regression tasks in non-NLP domains. We outperform task-specific specialized models on 3 of 4 tasks and almost match them on binding affinity. Note that Text+Chem T5 and LlaSMol are chemistry-specific models, so we only evaluate them on chemistry-related tasks but not protein-related ones.}
\label{table:proteindrug}
\vspace{-1mm}
\renewcommand{\arraystretch}{1.1}
\resizebox{\textwidth}{!}{
\begin{tabular}{lcccccc}
		\toprule
  && & \multicolumn{2}{c}{Single-Domain, Single-Instance} & Single-Domain, Multi-Instance & Multi-Domain\\
   \cmidrule(lr){4-5} 
\cmidrule(lr){6-7} 
  & Fine-Tuning &\# Trainable & Descriptor       &  QED  & Drug Combination  & Binding Affinity \\  
 \cmidrule(lr){4-4}  
 \cmidrule(lr){5-5} \cmidrule(lr){6-6}    \cmidrule(lr){7-7}
Model & Method& Params & MSE ($\downarrow$)  & MSE ($\downarrow$)  & MAE ($\downarrow$)& Pearson $r$ ($\uparrow$)\\ 
\midrule
\textbf{Specialized}&\\
{Nearest Neighbor}&&0&0.012&0.040&-&-\\
{Ensemble Model}&&9M&-&-&-&\textbf{0.588}\\
{PLM}&&4M&-&-&-&0.538\\
{MLP}&&13M&-&-&16.85&0.433\\
 \midrule
  \textbf{Direct Prompting}&\\
  {LLaMA}&&0 &0.236&0.298&23.39&0.017\\
 GPT-4	&&0&0.031&0.18&	22.35	&	-0.1051\\
Galactica &&0&	0.089 &0.36	&22.4	&	-0.0053\\
Text+Chem T5 &&0&-&0.31&	26.60&-\\
LlaSMol&&0&-&0.061	&24.15&-\\
\midrule
	\textbf{Fine-Tuning Based}\\

Galactica & Linear Probing &2K& 0.022 &0.13&	22.76		&0.1557\\
Text+Chem T5 & Linear Probing&768	&-&0.048	&22.22&-\\
LlaSMol & Linear Probing&4K&-&	0.016	&21.43&-\\
 LLaMA & Linear Probing&4K & 0.041&0.012&24.11 &0.180\\
LLaMA & Prompt Tuning&86K&0.011&1.387&17.65&0.054	  \\
 {LLaMA }&LoRA&1M&0.006&0.020&13.46&0.125\\
{Tag-LLaMA \textbf{(Ours)}}& Tag-LLM &86K&\textbf{0.005}&\textbf{0.008}	&\textbf{12.21}&0.527\\

\bottomrule
\end{tabular}}
\vspace{-4mm}
\end{table*}
\section{Repurposing LLMs for Scientific Discovery}
\label{sec:exp:science}

For scientific domains, developing large specialized models can be costly due to limited data and the need for domain expertise. In this section, we study how LLMs can be applied to two such domains: proteins (or amino acid sequences) and chemical compounds, which are represented by the simplified molecular-input line-entry system (SMILES).

In biomedical research, a protein is usually represented as a string of letters (e.g., \texttt{MTVPDRSEIAG...}) where either a single- or three-letter code represents an amino acid, ordered from the amino-terminal to the carboxyl-terminal of the protein. Similarly, SMILES is a notation system for describing the structure of chemical compounds, e.g., \texttt{CC(=O)NC1=CC=C(C=C1)O} for Tylenol. Following our 3-stage protocol, we first train the \token{Protein} and \token{SMILES} tags for next-token prediction
using unlabeled data extracted from \citet{badataset}. 
Previous work has shown that general LLMs on their own have poor understanding of these special notations \citep{evidence1, evidence2}. Nonetheless, we will show that our method enables LLMs to process these representations and make predictions with remarkable precision---it not only outperforms prompt tuning and fine-tuning baselines by a large margin, but also performs competitively against  highly specialized expert models in a variety of protein-related and drug discovery  tasks. 

\subsection{Single-Domain, Single-Instance Tasks} 
\label{sec:singledomain_singleinstance}

We start with two single-domain, single-instance tasks:  descriptor prediction for proteins, and Quantative Estimation of Drug-likeness (QED) score prediction for chemical compounds. The ground truth labels are automatically generated using the Python \href{https://peptides.readthedocs.io/en/stable/index.html}{peptides} and \href{https://www.rdkit.org/docs/GettingStartedInPython.html}{RDKit} libraries, respectively. Since the tasks involve numerical prediction, we train the function tags and the paired regression heads using the MSE loss. We compare with three sets of baselines. First, we directly prompt state-of-the-art LLMs with task instructions. We consider five models: the general-purpose GPT-4 \citep{GPT4} and LLaMA-7B, the science-focused Galactica-1.3B \citep{GALACTICA}, the chemistry-specialized LlaSMol \citep{yu2024llasmol} and Text+Chem T5-Base \citep{chemt5}. Second, we apply PEFT methods including LoRA \citep{lora}, prompt tuning \citep{prompttuning}, and linear probing to LLaMA-7B. We also apply linear probing to the specialized LLMs mentioned earlier. Lastly, we consider task-specific, non-LLM models and non-parametric methods, such as the nearest-neighbor algorithm that looks for similar training data points and leverages their labels to make predictions. For more details about the tasks and baselines, see Appendix~\ref{appendix:sec:baseline}.

We report test MSE and Pearson's correlation coefficient in Table~\ref{table:proteindrug}. Notably, our  method obtains the lowest error rates and the highest correlation scores among all baselines on both tasks.
Table~\ref{table:proteindrug} also compares the number of learnable parameters for each approach. Our method is much more parameter-efficient than the second-ranked LoRA, and much more effective than vanilla prompt tuning, which uses  the same number of parameters as we do.
Encouraged by these  results, we turn to more challenging tasks that involve multi-instance prediction for drug discovery.

\begin{table*}
\Large
\centering
  \caption{\small Augmenting PEFT baselines with  domain name or regression head  as described in Section~\ref{sec:ablation}.  Our method outperforms naive prompt tuning on all tasks and outperforms all baselines on 3 of 4 tasks. Note that the regression head is also used in this setting.
}
\label{table:reghead_domain}
\vspace{-2mm}
\renewcommand{\arraystretch}{1.1}
\resizebox{\textwidth}{!}{
\begin{tabular}{llccccc}
		\toprule
   
 & & \# Trainable Params&Descriptor  (MSE, $\downarrow$)  & QED  (MSE, $\downarrow$) &DC  (MAE, $\downarrow$)  &BA (Pearson $r$, $\uparrow$)\\  
\midrule
\Algo&&86K&\textbf{0.005}&\textbf{0.008}&\textbf{12.21}&0.527	\\
\midrule
{LoRA}&{+ Text-Form Domain Info}&1M& 0.008&0.015&12.79&\textbf{0.552}\\
Prompt Tuning&+ Text-Form Domain Info&86K& 0.007&0.0085&14.73	  &	0.37\\
Linear Probing&+ Text-Form Domain Info &4K&  0.049& 0.011&14.64&0.31\\
\midrule
{LoRA}&{ + Regression Head}&1M& 0.007&0.015&12.53&0.534\\
Prompt Tuning&{ + Regression Head}&86K& 0.008&0.0083&15.29	  &	0.38\\
Linear Probing&{ + Regression Head}&4K&  0.041& 0.012&24.11&0.18\\
\bottomrule
	\end{tabular}}
\end{table*}

\subsection{Single-Domain, Multi-Instance Tasks} 
\label{sec:singledomain_multiinstance}

The Therapeutics Data Commons (TDC)  \citep{TDC} covers a wide range of tasks in therapeutic modalities related to target discovery and activity modeling. 
We start with the Drug Combination (DC) task that predicts the sensitivity score between a pair of drugs, using the DrugComb\_CSS benchmark. To preprocess the data, we insert \token{SMILES} before the SMILES strings for the drug compounds and append \token{DC} to the end of the input. We learn these tags as well as the regression head using the MSE loss.

Following the benchmark, we compute the mean-absolute-error (MAE) on the test set (Table~\ref{table:proteindrug}). Our method achieves state-of-the-art on drug combination. It is worth noting that we outperform not only all LLM-based baselines but also the domain-specific expert model \citep{dcexpert} trained with supervised in-domain data. This shows the benefit of leveraging LLMs, since these pretrained LLMs are both large in scale and equipped with general knowledge.

\subsection{Multi-Domain, Multi-Instance Tasks}
\label{sec:multidomain_multiinstance}

The Drug Combination task still operates within a single SMILES domain. Hence, to complete our evaluation, we study a cross-domain task---predicting the binding affinity between a small-molecule drug and a target protein. Traditional methods to gauge the affinities require  expensive wet-lab experiments, limiting the number of candidate drugs that researchers can search over. However, if we can repurpose LLMs for predicting binding affinities,
we can not only reduce the pharmaceutical research costs but also  enlarge the search space to avoid missing potential candidates.

To see how our method performs on this challenging task, we use TDC's BindingDB\_Patent benchmark. The training and test sets  are split by the label's patent year to simulate distribution shift. We report the Pearson $r$  in Table~\ref{table:proteindrug}. Once again, our method outperforms all LLM-based baselines, providing evidence that the learned input tags are robust under distribution shifts. However, we rank the third among all models submitted to the leaderboard, underperforming the ensemble-based Otter-Knowledge \citep{baexpert1} and the PLM model \citep{baexpert2}, all of which are highly specialized methods based on pretrained protein representations. The small performance gap shows the potential of using pretrained LLMs for scientific problems.
We believe that an interesting future direction of this work is to explore whether scaling from LLaMA-7B to larger models can result in better performance than these specialized methods.

\section{Ablation Studies: Understand \Algo} 
\label{sec:ablation}
In this section, we provide a detailed study of our method and the three-stage training protocol using the scientific tasks. We show that learning and using input tags significantly improves the LLM's performance on specialized tasks over naive prompt tuning and is much more parameter-efficient than adapter-based methods like LoRA.

\subsection{Ablations on Input Tags}
\label{sec:ablation_ability}

We first study the effect of input tags using the DC dataset. We separately examine domain tags and function tags by removing one type of tags from the input prompt and then adapt the LLM.
Table~\ref{fig:ablation_setting} shows that removing any input tags leads to worse MAE compared to the complete workflow. Moreover, removing the function tags results in a more significant  performance drop, which implies that allocating trainable parameters for extracting task-relevant knowledge is necessary.
To further explore how  the domain tags benefit adaptation, we perform three sets experiments. 

\textbf{Use of Auxiliary Data } First, since the domain tags are trained for next-token generation using auxiliary data, they should extract external knowledge unknown to the LLM, such as the conditional distribution of specialized domains. 
To study whether such knowledge is helpful,
we replace the learned embeddings of the domain tags with 
actual text tokens ``\texttt{$<$}", ``\texttt{Protein}", ``\texttt{$>$}"
and fine-tune the LLM using PEFT methods.
As shown in the ``\textsc{domain in text}" rows of Table~\ref{table:reghead_domain}, simply specifying the domain as text brings marginal benefits  to prompt tuning and LoRA. 
This shows the importance of incorporating external data, as we do in stage 1. Our method also outperforms these baselines on all tasks except for BA.
Though LoRA slightly outperforms our method  by $0.018$ on BA,  it uses 12x more parameters than we do ($\sim$1M vs. $\sim$86K).

\looseness=-1
The above experiment also suggests that using learnable tags enables more fine-grained control over how we condition the LLM. Specifically, the effect of using actual text like  ``\texttt{Protein}" to condition the model relies heavily on its occurrence in the pretraining corpora, but end users do not have control over this---we are uncertain how frequently the word ``\texttt{Protein}" occurs and whether these occurrences align with the task of interest.
Our method addresses this limitation by learning the tag embeddings explicitly from the target domain's data. 

\textbf{Effect of enrichment.} Second, recall that our training protocol specifies how domain tags can be enriched in stage 2  (Section~\ref{sec:training:stage1}). We verify that enrichment is indeed beneficial---Table~\ref{fig:ablation_setting} shows that using the \token{SMILES} tag enriched with QED knowledge achieves a even lower error rate than directly using the non-enriched tag.

\begin{figure}
\vspace{-2mm}
    \centering
    \includegraphics[width=0.42\textwidth]{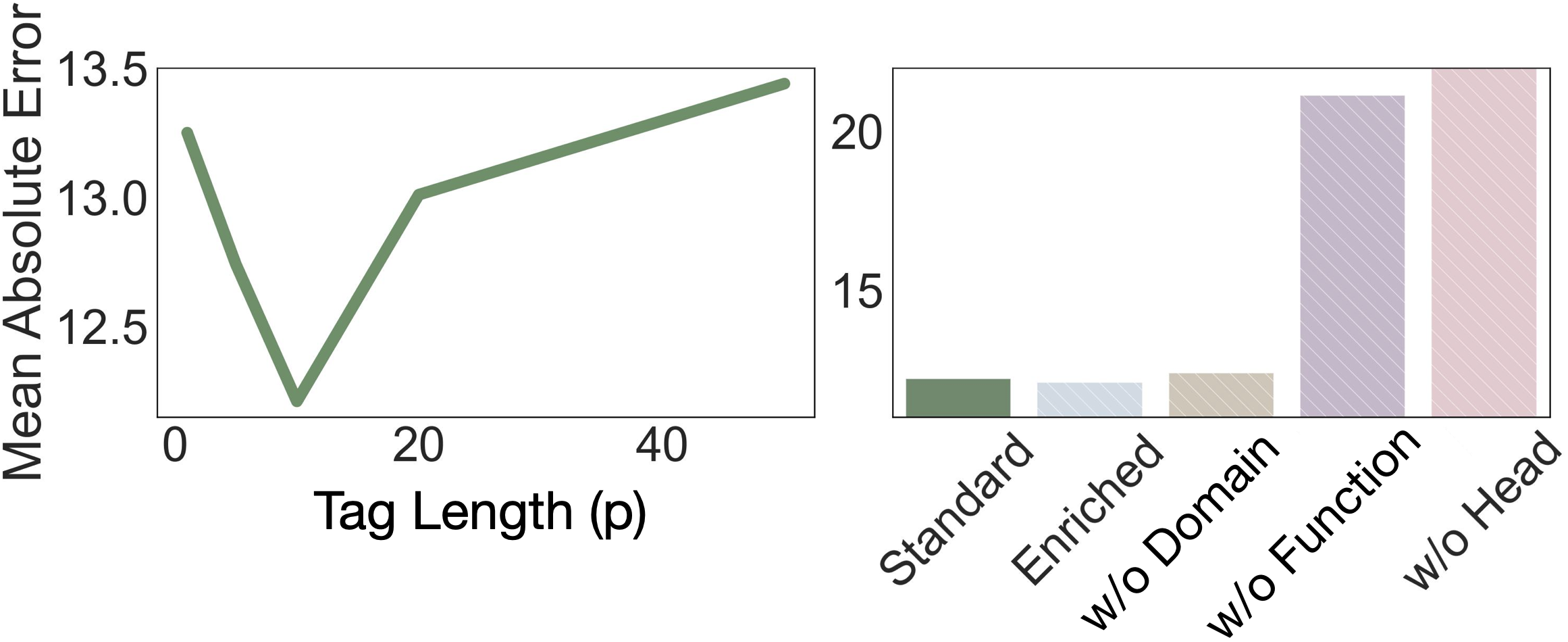}
    \vspace{-3mm}
    \caption{\small  \textbf{Left:} MAE for varying lengths of the \token{DC} tag. As $p$ increases, the test error first decreases and then rises. The empirical optimum is $p=10$. \textbf{Right:} visualizing ablation performance.}
    \label{fig:ablation_tokenlen}
    \vspace{-3mm}
\end{figure}

\textbf{Effect of Tag Length } Lastly, we study the effect of $p$, the number of tokens used for representing each tag. We vary the length of the \token{DC} tag, taking $p\in\{1, 5, 10, 20, 50\}$. As shown in Figure~\ref{fig:ablation_tokenlen}, as $p$ becomes larger, the test error first decreases and then increases. This suggests that, while the added degrees of freedom are initially beneficial, going beyond a certain threshold might result in overfitting to the training data, thus hindering  test-time performance. 

\begin{table}
\Large
\vspace{-2mm}

\caption{\small Enriching domain tag improves performance, whereas removing the tags or regression head hurts performance. }
\vspace{-2mm}
\label{fig:ablation_setting}
    \resizebox{0.48\textwidth}{!}{
\begin{tabular}{ccccc}
			\toprule
			   &Domain Tags & Function Tags &Regression Head & MAE($\downarrow$)   \\
			\midrule
	Full  &\checkmark&\checkmark&\checkmark&12.21\\
 Enriched & \checkmark&\checkmark&\checkmark&12.10\\	
w/o Domain Tags&&\checkmark&\checkmark&12.39\\
w/o Function Tags&\checkmark&&\checkmark&21.14\\	
w/o Reg Head &\checkmark&\checkmark&&23.42\\
\bottomrule
		\end{tabular}}
  \vspace{-3mm}
  \end{table}
\subsection{Ablations on Regression Head} 
\label{sec:ablation_regression}
We also study how the regression head affects our method by disabling it---instead of making scalar predictions, we  generate digits one by one via the language modeling head. As shown in Table~\ref{fig:ablation_setting}, the ``w/o Reg Head" setting results in a notable performance drop  on the DC dataset. This empirically validates our hypothesis in Section~\ref{sec:beyond_text} that next-token prediction is ill-suited for regression tasks.

To further verify this idea, we augment the PEFT baselines with the regression head and train them using the MSE loss (Table~\ref{table:reghead_domain}). This improves the performance for all baselines compared to digit generation. 
The regression-augmented PEFT baselines also helps us understand the effect of input tags, since  they remove the effect of the prediction head and the loss function. As shown in Table~\ref{table:reghead_domain}, our method outperforms prompt tuning on all tasks and outperforms all regression-augmented PEFT methods
on 3 of 4 tasks.
This suggests that our way of learning and using input tags  is effective for improving downstream performance. The gap between our method and LoRA on binding affinity is likely due to the fact that we use much fewer trainable parameters. 

In all of the above experiments, we use LLaMA-7B as the LLM backbone to showcase the efficacy of our method. In Appendix~\ref{appendix:mistral},  we further investigate applying \Algo to another open-source model, Mistral-7B~\citep{jiang2023mistral}. The results demonstrate the broad applicability of \Algo beyond just the LLaMA models.

\section{Conclusion and Future Work}
In this work, we exploit off-the-shelf LLMs for solving specialized tasks. We developed a system of LLM tags that condition the LLM and proposed a three-stage training protocol to learn 
the tags. Our experimental results show that our method improves the LLM's prediction quality and allows for more fine-grained control over its behavior. We envision open-sourcing the learned tags for different models can help facilitate research in specialized domains.

We identify several future directions based on our work. For example, we can further validate our method in other specialized domains, such as gene function prediction (computational biology) or solving partial differential equations (physics).
The idea of enhancing function tags with task-specific output heads can be applied to various prediction problems. However, in this study, we mainly focused on regression, leaving the exploration of classification and other structured prediction problems
for future research. In terms of computational efficiency, a potential improvement could come from training the tags in large batches, e.g., by concatenating data from different domains together, instead of sequentially as we do here. Lastly, integrating our work with other adaptation paradigms, such as in-context learning, presents an intriguing possibility for exploration.


\section*{Impact Statement}

This paper calls for ML community's attention to take advantage of  LLMs and apply them to a wider range of real-world problems beyond the NLP domains. This moves towards truly democratizing machine learning in real life. In terms of broader societal impact, our work can exert a positive influence as it  contributes to reusing existing models and resources, reducing the computational burden of developing new large-scale models on massive data. However, lowering the barrier for applying LLMs to a wide range of tasks necessarily comes with the risk of misuse. Hence, it is imperative to develop  adaptation methods with  better privacy, safety, and fairness guarantees.

\nocite{langley00}

\bibliography{example_paper}

\begin{thebibliography}{53}
\providecommand{\natexlab}[1]{#1}
\providecommand{\url}[1]{\texttt{#1}}
\expandafter\ifx\csname urlstyle\endcsname\relax
  \providecommand{\doi}[1]{doi: #1}\else
  \providecommand{\doi}{doi: \begingroup \urlstyle{rm}\Url}\fi

\bibitem[Ahmad et~al.(2020)Ahmad, Chakraborty, Ray, and Chang]{code2}
Ahmad, W.~U., Chakraborty, S., Ray, B., and Chang, K.-W.
\newblock A transformer-based approach for source code summarization.
\newblock \emph{ArXiv}, abs/2005.00653, 2020.

\bibitem[Bickerton et~al.(2012)Bickerton, Paolini, Besnard, Muresan, and Hopkins]{QED}
Bickerton, G. R.~J., Paolini, G.~V., Besnard, J., Muresan, S., and Hopkins, A.~L.
\newblock Quantifying the chemical beauty of drugs.
\newblock \emph{Nature chemistry}, 4 2:\penalty0 90--8, 2012.

\bibitem[Blanchard et~al.(2021)Blanchard, Gounley, Bhowmik, Shekar, Lyngaas, Gao, Yin, Tsaris, Wang, and Glaser]{badataset}
Blanchard, A.~E., Gounley, J.~P., Bhowmik, D., Shekar, M.~C., Lyngaas, I., Gao, S., Yin, J., Tsaris, A., Wang, F., and Glaser, J.
\newblock Language models for the prediction of sars-cov-2 inhibitors.
\newblock \emph{The International Journal of High Performance Computing Applications}, 36:\penalty0 587 -- 602, 2021.

\bibitem[Brown et~al.(2020)Brown, Mann, Ryder, Subbiah, Kaplan, Dhariwal, Neelakantan, Shyam, Sastry, Askell, Agarwal, Herbert-Voss, Krueger, Henighan, Child, Ramesh, Ziegler, Wu, Winter, Hesse, Chen, Sigler, Litwin, Gray, Chess, Clark, Berner, McCandlish, Radford, Sutskever, and Amodei]{Brown2020Language}
Brown, T.~B., Mann, B., Ryder, N., Subbiah, M., Kaplan, J., Dhariwal, P., Neelakantan, A., Shyam, P., Sastry, G., Askell, A., Agarwal, S., Herbert-Voss, A., Krueger, G., Henighan, T.~J., Child, R., Ramesh, A., Ziegler, D.~M., Wu, J., Winter, C., Hesse, C., Chen, M., Sigler, E., Litwin, M., Gray, S., Chess, B., Clark, J., Berner, C., McCandlish, S., Radford, A., Sutskever, I., and Amodei, D.
\newblock Language models are few-shot learners.
\newblock \emph{ArXiv}, abs/2005.14165, 2020.

\bibitem[Chan et~al.(2020)Chan, Ong, Pung, Zhang, and Fu]{generation2}
Chan, A., Ong, Y., Pung, B. T.~W., Zhang, A., and Fu, J.
\newblock Cocon: A self-supervised approach for controlled text generation.
\newblock \emph{ArXiv}, abs/2006.03535, 2020.

\bibitem[Chilingaryan et~al.(2022)Chilingaryan, Tamoyan, Tevosyan, Babayan, Khondkaryan, Hambardzumyan, Navoyan, Khachatrian, and Aghajanyan]{Chilingaryan2022BARTSmilesGM}
Chilingaryan, G., Tamoyan, H., Tevosyan, A., Babayan, N., Khondkaryan, L., Hambardzumyan, K., Navoyan, Z., Khachatrian, H., and Aghajanyan, A.
\newblock Bartsmiles: Generative masked language models for molecular representations.
\newblock \emph{ArXiv}, abs/2211.16349, 2022.

\bibitem[Christofidellis et~al.(2023)Christofidellis, Giannone, Born, Winther, Laino, and Manica]{chemt5}
Christofidellis, D., Giannone, G., Born, J., Winther, O., Laino, T., and Manica, M.
\newblock Unifying molecular and textual representations via multi-task language modelling.
\newblock 202:\penalty0 6140--6157, 2023.
\newblock URL \url{https://proceedings.mlr.press/v202/christofidellis23a.html}.

\bibitem[Dathathri et~al.(2019)Dathathri, Madotto, Lan, Hung, Frank, Molino, Yosinski, and Liu]{generation1}
Dathathri, S., Madotto, A., Lan, J., Hung, J., Frank, E., Molino, P., Yosinski, J., and Liu, R.
\newblock Plug and play language models: A simple approach to controlled text generation.
\newblock \emph{ArXiv}, abs/1912.02164, 2019.

\bibitem[Dinh et~al.(2022)Dinh, Zeng, Zhang, Lin, Rajput, Gira, yong Sohn, Papailiopoulos, and Lee]{LIFT}
Dinh, T., Zeng, Y., Zhang, R., Lin, Z., Rajput, S., Gira, M., yong Sohn, J., Papailiopoulos, D., and Lee, K.
\newblock Lift: Language-interfaced fine-tuning for non-language machine learning tasks.
\newblock \emph{ArXiv}, abs/2206.06565, 2022.

\bibitem[Fan et~al.(2020)Fan, Bhosale, Schwenk, Ma, El-Kishky, Goyal, Baines, Çelebi, Wenzek, Chaudhary, Goyal, Birch, Liptchinsky, Edunov, Grave, Auli, and Joulin]{m2m}
Fan, A., Bhosale, S., Schwenk, H., Ma, Z., El-Kishky, A., Goyal, S., Baines, M., Çelebi, O., Wenzek, G., Chaudhary, V., Goyal, N., Birch, T., Liptchinsky, V., Edunov, S., Grave, E., Auli, M., and Joulin, A.
\newblock Beyond english-centric multilingual machine translation.
\newblock \emph{ArXiv}, abs/2010.11125, 2020.

\bibitem[Georgiev(2009)]{descriptor}
Georgiev, A.~G.
\newblock Interpretable numerical descriptors of amino acid space.
\newblock \emph{Journal of computational biology : a journal of computational molecular cell biology}, 16 5:\penalty0 703--23, 2009.

\bibitem[Goyal et~al.(2021)Goyal, Gao, Chaudhary, Chen, Wenzek, Ju, Krishnan, Ranzato, Guzm{\'a}n, and Fan]{flores}
Goyal, N., Gao, C., Chaudhary, V., Chen, P.-J., Wenzek, G., Ju, D., Krishnan, S., Ranzato, M., Guzm{\'a}n, F., and Fan, A.
\newblock The flores-101 evaluation benchmark for low-resource and multilingual machine translation.
\newblock \emph{Transactions of the Association for Computational Linguistics}, 10:\penalty0 522--538, 2021.

\bibitem[Guo et~al.(2023)Guo, Guo, Liang, Guo, Chawla, Wiest, Zhang, et~al.]{guo2023indeed}
Guo, T., Guo, K., Liang, Z., Guo, Z., Chawla, N.~V., Wiest, O., Zhang, X., et~al.
\newblock What indeed can gpt models do in chemistry? a comprehensive benchmark on eight tasks.
\newblock \emph{arXiv preprint arXiv:2305.18365}, 2023.

\bibitem[Hendrycks et~al.(2021)Hendrycks, Burns, Kadavath, Arora, Basart, Tang, Song, and Steinhardt]{hendrycks2021measuring}
Hendrycks, D., Burns, C., Kadavath, S., Arora, A., Basart, S., Tang, E., Song, D., and Steinhardt, J.
\newblock Measuring mathematical problem solving with the math dataset.
\newblock In \emph{Thirty-fifth Conference on Neural Information Processing Systems Datasets and Benchmarks Track (Round 2)}, 2021.

\bibitem[Hou \& Ji(2023)Hou and Ji]{evidence3}
Hou, W. and Ji, Z.
\newblock Geneturing tests gpt models in genomics.
\newblock \emph{bioRxiv}, 2023.

\bibitem[Hu et~al.(2021)Hu, Shen, Wallis, Allen-Zhu, Li, Wang, and Chen]{lora}
Hu, J.~E., Shen, Y., Wallis, P., Allen-Zhu, Z., Li, Y., Wang, S., and Chen, W.
\newblock Lora: Low-rank adaptation of large language models.
\newblock \emph{ArXiv}, abs/2106.09685, 2021.

\bibitem[Huang et~al.(2021)Huang, Fu, Gao, Zhao, Roohani, Leskovec, Coley, Xiao, Sun, and Zitnik]{TDC}
Huang, K., Fu, T., Gao, W., Zhao, Y., Roohani, Y., Leskovec, J., Coley, C.~W., Xiao, C., Sun, J., and Zitnik, M.
\newblock Therapeutics data commons: Machine learning datasets and tasks for drug discovery and development.
\newblock \emph{Proceedings of Neural Information Processing Systems, NeurIPS Datasets and Benchmarks}, 2021.

\bibitem[Jiang et~al.(2023)Jiang, Sablayrolles, Mensch, Bamford, Chaplot, de~las Casas, Bressand, Lengyel, Lample, Saulnier, Lavaud, Lachaux, Stock, Scao, Lavril, Wang, Lacroix, and Sayed]{jiang2023mistral}
Jiang, A.~Q., Sablayrolles, A., Mensch, A., Bamford, C., Chaplot, D.~S., de~las Casas, D., Bressand, F., Lengyel, G., Lample, G., Saulnier, L., Lavaud, L.~R., Lachaux, M.-A., Stock, P., Scao, T.~L., Lavril, T., Wang, T., Lacroix, T., and Sayed, W.~E.
\newblock Mistral 7b, 2023.

\bibitem[Jumper et~al.(2021)Jumper, Evans, Pritzel, Green, Figurnov, Ronneberger, Tunyasuvunakool, Bates, Z{\'i}dek, Potapenko, Bridgland, Meyer, Kohl, Ballard, Cowie, Romera-Paredes, Nikolov, Jain, Adler, Back, Petersen, Reiman, Clancy, Zielinski, Steinegger, Pacholska, Berghammer, Bodenstein, Silver, Vinyals, Senior, Kavukcuoglu, Kohli, and Hassabis]{alphafold}
Jumper, J.~M., Evans, R., Pritzel, A., Green, T., Figurnov, M., Ronneberger, O., Tunyasuvunakool, K., Bates, R., Z{\'i}dek, A., Potapenko, A., Bridgland, A., Meyer, C., Kohl, S. A.~A., Ballard, A., Cowie, A., Romera-Paredes, B., Nikolov, S., Jain, R., Adler, J., Back, T., Petersen, S., Reiman, D.~A., Clancy, E., Zielinski, M., Steinegger, M., Pacholska, M., Berghammer, T., Bodenstein, S., Silver, D., Vinyals, O., Senior, A.~W., Kavukcuoglu, K., Kohli, P., and Hassabis, D.
\newblock Highly accurate protein structure prediction with alphafold.
\newblock \emph{Nature}, 596:\penalty0 583 -- 589, 2021.

\bibitem[Keskar et~al.(2019)Keskar, McCann, Varshney, Xiong, and Socher]{CTRL}
Keskar, N.~S., McCann, B., Varshney, L.~R., Xiong, C., and Socher, R.
\newblock Ctrl: A conditional transformer language model for controllable generation.
\newblock \emph{ArXiv}, abs/1909.05858, 2019.

\bibitem[Lam et~al.(2023)Lam, Sbodio, Galindo, Zayats, Fern'andez-D'iaz, Valls, Picco, Ramis, and L'opez]{baexpert1}
Lam, H.~T., Sbodio, M.~L., Galindo, M.~M., Zayats, M., Fern'andez-D'iaz, R., Valls, V., Picco, G., Ramis, C.~B., and L'opez, V.
\newblock Otter-knowledge: benchmarks of multimodal knowledge graph representation learning from different sources for drug discovery.
\newblock \emph{ArXiv}, abs/2306.12802, 2023.

\bibitem[Lester et~al.(2021)Lester, Al{-}Rfou, and Constant]{prompttuning}
Lester, B., Al{-}Rfou, R., and Constant, N.
\newblock The power of scale for parameter-efficient prompt tuning.
\newblock \emph{CoRR}, abs/2104.08691, 2021.
\newblock URL \url{https://arxiv.org/abs/2104.08691}.

\bibitem[Lewkowycz et~al.(2022)Lewkowycz, Andreassen, Dohan, Dyer, Michalewski, Ramasesh, Slone, Anil, Schlag, Gutman-Solo, Wu, Neyshabur, Gur-Ari, and Misra]{quantitative}
Lewkowycz, A., Andreassen, A., Dohan, D., Dyer, E., Michalewski, H., Ramasesh, V.~V., Slone, A., Anil, C., Schlag, I., Gutman-Solo, T., Wu, Y., Neyshabur, B., Gur-Ari, G., and Misra, V.
\newblock Solving quantitative reasoning problems with language models.
\newblock \emph{ArXiv}, abs/2206.14858, 2022.

\bibitem[Li \& Liang(2021)Li and Liang]{prefixtuning}
Li, X.~L. and Liang, P.
\newblock Prefix-tuning: Optimizing continuous prompts for generation.
\newblock In \emph{Proceedings of the 59th Annual Meeting of the Association for Computational Linguistics and the 11th International Joint Conference on Natural Language Processing (Volume 1: Long Papers)}, pp.\  4582--4597, Online, August 2021. Association for Computational Linguistics.
\newblock \doi{10.18653/v1/2021.acl-long.353}.
\newblock URL \url{https://aclanthology.org/2021.acl-long.353}.

\bibitem[Lin et~al.(2021)Lin, Mihaylov, Artetxe, Wang, Chen, Simig, Ott, Goyal, Bhosale, Du, Pasunuru, Shleifer, Koura, Chaudhary, O'Horo, Wang, Zettlemoyer, Kozareva, Diab, Stoyanov, and Li]{xglm}
Lin, X.~V., Mihaylov, T., Artetxe, M., Wang, T., Chen, S., Simig, D., Ott, M., Goyal, N., Bhosale, S., Du, J., Pasunuru, R., Shleifer, S., Koura, P.~S., Chaudhary, V., O'Horo, B., Wang, J., Zettlemoyer, L., Kozareva, Z., Diab, M.~T., Stoyanov, V., and Li, X.
\newblock Few-shot learning with multilingual generative language models.
\newblock In \emph{Conference on Empirical Methods in Natural Language Processing}, 2021.

\bibitem[Lu et~al.(2022{\natexlab{a}})Lu, Grover, Abbeel, and Mordatch]{FPT}
Lu, K., Grover, A., Abbeel, P., and Mordatch, I.
\newblock Frozen pretrained transformers as universal computation engines.
\newblock \emph{Proceedings of the AAAI Conference on Artificial Intelligence}, 36\penalty0 (7):\penalty0 7628--7636, Jun. 2022{\natexlab{a}}.

\bibitem[Lu et~al.(2022{\natexlab{b}})Lu, Bartolo, Moore, Riedel, and Stenetorp]{lu2022fantastically}
Lu, Y., Bartolo, M., Moore, A., Riedel, S., and Stenetorp, P.
\newblock Fantastically ordered prompts and where to find them: Overcoming {Few-Shot} prompt order sensitivity.
\newblock In \emph{Proceedings of the 60th Annual Meeting of the Association for Computational Linguistics (Volume 1: Long Papers)}, pp.\  8086--8098, Dublin, Ireland, May 2022{\natexlab{b}}. Association for Computational Linguistics.
\newblock \doi{10.18653/v1/2022.acl-long.556}.

\bibitem[Mu et~al.(2023)Mu, Li, and Goodman]{gist}
Mu, J., Li, X.~L., and Goodman, N.~D.
\newblock Learning to compress prompts with gist tokens.
\newblock \emph{ArXiv}, abs/2304.08467, 2023.

\bibitem[OpenAI(2023)]{GPT4}
OpenAI.
\newblock Gpt-4 technical report.
\newblock 2023.

\bibitem[Qian et~al.(2022)Qian, Dong, Shen, Wei, and Chen]{prefixmarker}
Qian, J., Dong, L., Shen, Y., Wei, F., and Chen, W.
\newblock Controllable natural language generation with contrastive prefixes.
\newblock In \emph{Findings}, 2022.

\bibitem[Roberts et~al.(2021)Roberts, Guo, Xu, Talwalkar, Lander, Tao, Cai, Niu, Heng, Qin, Deng, Hog, Pfefferle, Shivakumar, Krishnakumar, Wang, Sukthanker, Hutter, Hasanaj, Le, Khodak, Nevmyvaka, Rasul, Sala, Schneider, Shen, and Sparks]{Roberts2021AutoMLDD}
Roberts, N., Guo, S., Xu, C., Talwalkar, A., Lander, D., Tao, L., Cai, L., Niu, S., Heng, J., Qin, H., Deng, M., Hog, J., Pfefferle, A., Shivakumar, S.~A., Krishnakumar, A., Wang, Y., Sukthanker, R.~S., Hutter, F., Hasanaj, E., Le, T.-D., Khodak, M., Nevmyvaka, Y., Rasul, K., Sala, F., Schneider, A., Shen, J., and Sparks, E.~R.
\newblock Automl decathlon: Diverse tasks, modern methods, and efficiency at scale.
\newblock In \emph{Neural Information Processing Systems}, 2021.
\newblock URL \url{https://api.semanticscholar.org/CorpusID:265536645}.

\bibitem[Sanh et~al.(2021)Sanh, Webson, Raffel, Bach, Sutawika, Alyafeai, Chaffin, Stiegler, Scao, Raja, Dey, Bari, Xu, Thakker, Sharma, Szczechla, Kim, Chhablani, Nayak, Datta, Chang, Jiang, Wang, Manica, Shen, Yong, Pandey, Bawden, Wang, Neeraj, Rozen, Sharma, Santilli, F{\'e}vry, Fries, Teehan, Biderman, Gao, Bers, Wolf, and Rush]{T0}
Sanh, V., Webson, A., Raffel, C., Bach, S.~H., Sutawika, L., Alyafeai, Z., Chaffin, A., Stiegler, A., Scao, T.~L., Raja, A., Dey, M., Bari, M.~S., Xu, C., Thakker, U., Sharma, S., Szczechla, E., Kim, T., Chhablani, G., Nayak, N.~V., Datta, D., Chang, J.~D., Jiang, M. T.-J., Wang, H., Manica, M., Shen, S., Yong, Z.~X., Pandey, H., Bawden, R., Wang, T., Neeraj, T., Rozen, J., Sharma, A., Santilli, A., F{\'e}vry, T., Fries, J.~A., Teehan, R., Biderman, S.~R., Gao, L., Bers, T., Wolf, T., and Rush, A.~M.
\newblock Multitask prompted training enables zero-shot task generalization.
\newblock \emph{ArXiv}, abs/2110.08207, 2021.

\bibitem[Scao et~al.(2022)Scao, Fan, Akiki, Pavlick, Xie, Ye, Bras, Belkada, and Wolf]{bloom}
Scao, T.~L., Fan, A., Akiki, C., Pavlick, E.-J., Xie, .~Z., Ye, Z., Bras, M., Belkada, Y., and Wolf, T.
\newblock Bloom: A 176b-parameter open-access multilingual language model.
\newblock \emph{ArXiv}, abs/2211.05100, 2022.

\bibitem[Shao et~al.(2023)Shao, Cai, Xu, Liao, Zheng, and Yang]{composition}
Shao, N., Cai, Z., Xu, H., Liao, C., Zheng, Y., and Yang, Z.
\newblock Compositional task representations for large language models.
\newblock In \emph{International Conference on Learning Representations}, 2023.

\bibitem[Shen et~al.(2022)Shen, Khodak, and Talwalkar]{shen2022dash}
Shen, J., Khodak, M., and Talwalkar, A.
\newblock Efficient architecture search for diverse tasks.
\newblock In \emph{Advances in Neural Information Processing Systems (NeurIPS)}, 2022.

\bibitem[Shen et~al.(2023)Shen, Li, Dery, Staten, Khodak, Neubig, and Talwalkar]{orca}
Shen, J., Li, L., Dery, L.~M., Staten, C., Khodak, M., Neubig, G., and Talwalkar, A.
\newblock Cross-modal fine-tuning: align then refine.
\newblock In \emph{Proceedings of the 40th International Conference on Machine Learning}, 2023.

\bibitem[Shen et~al.(2024)Shen, Marwah, and Talwalkar]{shen2024ups}
Shen, J., Marwah, T., and Talwalkar, A.
\newblock Ups: Towards foundation models for pde solving via cross-modal adaptation, 2024.

\bibitem[Singhal et~al.(2022)Singhal, Azizi, Tu, Mahdavi, Wei, Chung, Scales, Tanwani, Cole-Lewis, Pfohl, Payne, Seneviratne, Gamble, Kelly, Scharli, Chowdhery, Mansfield, y~Arcas, Webster, Corrado, Matias, Chou, Gottweis, Tomavsev, Liu, Rajkomar, Barral, Semturs, Karthikesalingam, and Natarajan]{clinical}
Singhal, K., Azizi, S., Tu, T., Mahdavi, S., Wei, J., Chung, H.~W., Scales, N., Tanwani, A.~K., Cole-Lewis, H.~J., Pfohl, S.~J., Payne, P.~A., Seneviratne, M.~G., Gamble, P., Kelly, C., Scharli, N., Chowdhery, A., Mansfield, P.~A., y~Arcas, B.~A., Webster, D.~R., Corrado, G.~S., Matias, Y., Chou, K. H.-L., Gottweis, J., Tomavsev, N., Liu, Y., Rajkomar, A., Barral, J.~K., Semturs, C., Karthikesalingam, A., and Natarajan, V.
\newblock Large language models encode clinical knowledge.
\newblock \emph{Nature}, 620:\penalty0 172 -- 180, 2022.

\bibitem[Sledzieski(2022)]{baexpert2}
Sledzieski, S.
\newblock Adapting protein language models for rapid dti prediction.
\newblock \emph{bioRxiv}, 2022.

\bibitem[Taylor et~al.(2022)Taylor, Kardas, Cucurull, Scialom, Hartshorn, Saravia, Poulton, Kerkez, and Stojnic]{GALACTICA}
Taylor, R., Kardas, M., Cucurull, G., Scialom, T., Hartshorn, A., Saravia, E., Poulton, A., Kerkez, V., and Stojnic, R.
\newblock Galactica: A large language model for science.
\newblock 2022.

\bibitem[Touvron et~al.(2023)Touvron, Lavril, Izacard, Martinet, Lachaux, Lacroix, Rozi{\`e}re, Goyal, Hambro, Azhar, Rodriguez, Joulin, Grave, and Lample]{llama}
Touvron, H., Lavril, T., Izacard, G., Martinet, X., Lachaux, M.-A., Lacroix, T., Rozi{\`e}re, B., Goyal, N., Hambro, E., Azhar, F., Rodriguez, A., Joulin, A., Grave, E., and Lample, G.
\newblock Llama: Open and efficient foundation language models.
\newblock \emph{ArXiv}, abs/2302.13971, 2023.

\bibitem[Tu et~al.(2022)Tu, Roberts, Khodak, Shen, Sala, and Talwalkar]{nasbench360}
Tu, R., Roberts, N., Khodak, M., Shen, J., Sala, F., and Talwalkar, A.
\newblock {NAS}-bench-360: Benchmarking neural architecture search on diverse tasks.
\newblock In \emph{Advances in Neural Information Processing Systems (NeurIPS) Datasets and Benchmarks Track}, 2022.

\bibitem[Vinod et~al.(2023)Vinod, Chen, and Das]{Vinod2023ReprogrammingPL}
Vinod, R., Chen, P.-Y., and Das, P.
\newblock Reprogramming pretrained language models for protein sequence representation learning.
\newblock \emph{ArXiv}, abs/2301.02120, 2023.

\bibitem[Walker et~al.(2023)Walker, Ghani, Kuemmerli, Nebiker, M{\"u}ller, Raptis, and Staubli]{evidence1}
Walker, H.~L., Ghani, S., Kuemmerli, C., Nebiker, C.~A., M{\"u}ller, B.~P., Raptis, D.~A., and Staubli, S.~M.
\newblock Reliability of medical information provided by chatgpt: Assessment against clinical guidelines and patient information quality instrument.
\newblock \emph{J Med Internet Res}, 25:\penalty0 e47479, Jun 2023.
\newblock ISSN 1438-8871.
\newblock \doi{10.2196/47479}.
\newblock URL \url{https://www.jmir.org/2023/1/e47479}.

\bibitem[Wang et~al.(2023)Wang, Feng, and Wei]{evidence2}
Wang, R., Feng, H., and Wei, G.-W.
\newblock Chatbots in drug discovery: A case study on anti-cocaine addiction drug development with chatgpt.
\newblock \emph{ArXiv}, 2023.

\bibitem[Xia et~al.(2018)Xia, Shukla, Brettin, Garcia-Cardona, Cohn, Allen, Maslov, Holbeck, Doroshow, Evrard, Stahlberg, and Stevens]{dcexpert}
Xia, F., Shukla, M., Brettin, T.~S., Garcia-Cardona, C., Cohn, J.~D., Allen, J.~E., Maslov, S., Holbeck, S.~L., Doroshow, J.~H., Evrard, Y.~A., Stahlberg, E.~A., and Stevens, R.~L.
\newblock Predicting tumor cell line response to drug pairs with deep learning.
\newblock \emph{BMC Bioinformatics}, 19, 2018.

\bibitem[Yang et~al.(2022)Yang, Liu, Lei, Yang, Xue, Chen, and Xie]{softpromptmarker}
Yang, K., Liu, D., Lei, W., Yang, B., Xue, M., Chen, B., and Xie, J.
\newblock Tailor: A prompt-based approach to attribute-based controlled text generation.
\newblock \emph{ArXiv}, abs/2204.13362, 2022.

\bibitem[Yu et~al.(2024)Yu, Baker, Chen, Ning, and Sun]{yu2024llasmol}
Yu, B., Baker, F.~N., Chen, Z., Ning, X., and Sun, H.
\newblock Llasmol: Advancing large language models for chemistry with a large-scale, comprehensive, high-quality instruction tuning dataset.
\newblock \emph{arXiv preprint arXiv:2402.09391}, 2024.

\bibitem[Yu et~al.(2021)Yu, Sagae, and Yu]{prefixmarker2}
Yu, D., Sagae, K., and Yu, Z.
\newblock Attribute alignment: Controlling text generation from pre-trained language models.
\newblock \emph{ArXiv}, abs/2103.11070, 2021.

\bibitem[Zhang et~al.(2020)Zhang, Williams, Titov, and Sennrich]{opus}
Zhang, B., Williams, P., Titov, I., and Sennrich, R.
\newblock Improving massively multilingual neural machine translation and zero-shot translation.
\newblock \emph{ArXiv}, abs/2004.11867, 2020.

\bibitem[Zhang \& Song(2022)Zhang and Song]{hardpromptmarker}
Zhang, H. and Song, D.
\newblock Discup: Discriminator cooperative unlikelihood prompt-tuning for controllable text generation.
\newblock In \emph{Conference on Empirical Methods in Natural Language Processing}, 2022.

\bibitem[Zhang et~al.(2022)Zhang, Song, Li, Zhou, and Song]{generationsurvey}
Zhang, H., Song, H., Li, S., Zhou, M., and Song, D.
\newblock A survey of controllable text generation using transformer-based pre-trained language models.
\newblock \emph{ACM Computing Surveys}, 2022.

\bibitem[Zong \& Krishnamachari(2023)Zong and Krishnamachari]{math}
Zong, M.~L. and Krishnamachari, B.
\newblock Solving math word problems concerning systems of equations with gpt-3.
\newblock In \emph{AAAI Conference on Artificial Intelligence}, 2023.

\end{thebibliography}
\bibliographystyle{icml2024}

\newpage
\appendix
\onecolumn
\newpage
\appendix
\section{Appendix}

\subsection{Experiment details: our method}
\label{appendix:ours}

\subsubsection{General setup}
Our experiments leverage the LLaMA-7B \citep{llama} model as the backbone. We obtain the checkpoints from here: \href{https://huggingface.co/huggyllama/llama-7b/tree/main}{https://huggingface.co/huggyllama/llama-7b/tree/main}.
We run our workflow on a single NVIDIA A100 GPU, using FP16 training and evaluation. Below are the hyperparameters used for our experiments (excluding ablation studies).
\begin{itemize}
\item Tag length: 10
\item Batch size: 4
\item Gradient accumulation: 8
\item Optimizer: AdamW
\item Learning rate: 1E-4
\item Weight decay: 0
\item Learning rate scheduler: cosine (with warmup rate 0.03)
\end{itemize}

\subsubsection{Task-specific setup}
\label{appendix:sec:tasksetup}
Here we provide more information on the tasks used in our evaluation.

\paragraph{Translate}
For training the language tags and \token{Translate} tag, we use the OPUS-100 dataset \citep{opus}, which is an English-centric multilingual corpus covering 100 languages, i.e., all training pairs include English on either the source or target side. For evaluation, we use a subset of the FLORES-101 dataset \citep{flores} and report results for the devtest split. The baseline results are taken from \citet{xglm, bloom}.

\paragraph{Descriptor} 
\href{https://peptides.readthedocs.io/en/stable/index.html}{peptides} is a Python package that computes common descriptors for protein sequences. In our experiments, we first obtain the protein sequences from the \href{https://huggingface.co/datasets/jglaser/binding\_affinity}{binding\_affinity} dataset (dropping the other features). We then use peptides to compute the average BLOSUM indices for all the amino acids in the peptide and retain the first index as our target label.
BLOSUM indices were derived from physicochemical properties that have been subjected to a VARIMAX analysis \citep{descriptor}.

\paragraph{QED} Similar to descriptor, we first collect SMILES sequences from the \href{https://huggingface.co/datasets/jglaser/binding\_affinity}{binding\_affinity} dataset. Next, we make use of the 
\href{https://www.rdkit.org/docs/GettingStartedInPython.html}{rdkit.Chem.QED} module to compute the quantitative estimation of drug-likeness (QED) score. QED was initially introduced by \citet{QED}. The empirical rationale of this measure reflects the underlying distribution of molecular properties including molecular weight, logP, topological polar surface area, number of hydrogen bond donors and acceptors, the number of aromatic rings and rotatable bonds, and the presence of unwanted chemical functionalities.

\paragraph{Drug Combination} Drug combinations offer exciting new treatment opportunities to increase drug use and efficacy. For instance, simultaneously modulating multiple targets can address the issue of drug resistance seen in cancer treatments. However, experimentally searching the space of possible drug combinations is often infeasible. Thus, developing ML models for this purpose can be highly useful. In our experiments, we focus on predicting the synergy (the deviation of observed drug combination response from expected effects had non-interaction) using the TDC.DrugComb\_CSS dataset \citep{TDC}. The target value is derived using the relative IC50 values of compounds and the areas under dose-response curves.

\paragraph{Binding Affinity} This task requires predicting  the interaction activity score between a drug and a target protein using only the compound structural information and protein's amino acid sequence. It is practically meaningful in that identifying high-affinity compounds is the first crucial step for drug discovery.  In our experiments, we directly use the BindingDB datasets from TDC's \href{https://tdcommons.ai/benchmark/dti\_dg\_group/overview/}{dti\_dg\_group}.

Additionally, we use OPUS-100 to train the language tags and the \href{https://huggingface.co/datasets/jglaser/binding\_affinity}{binding\_affinity} dataset to train \token{Protein} and \token{SMILES}.
More details about data splits and processing for each task can be found in our supplementary code. 

\subsubsection{Input format}

In Table~\ref{appendix:table:format}, we summarize how we format the inputs of different tasks to query the language model. In the table, \token{input} indicates the place where we insert the raw input of a data point, whereas \token{output} indicates the label (or prediction value) of that data point. The other \token{}'s denote the input tags. For regression tasks (whose loss functions are MSE), we take the output hidden states of the function tags and pass them to the linear head.

Note that many LLM tokenizers automatically group several characters together for protein/SMILES sequences, introducing tokenization bias. Thus, we manually enforce every character to be a separate token in our workflow.

\begin{table*}[t]
\caption{Input format of all evaluated tasks.}
\label{appendix:table:format}
\centering
\resizebox{\textwidth}{!}{
\begin{tabular}{ccccl}
			\toprule
			Task  & Train Epochs &Loss & Eval Metric  &Input Format  \\
			\midrule
Translate&1&CE&ROUGE&\texttt{\# \# Input: \token{src\_lang} \token{input}}\\	
&&&COMET&\texttt{\# \# Output: \token{tgt\_lang} \token{Translate} \token{output} }\\

   Descriptor&4&MSE&MSE&\texttt{\# \# Input: The protein sequence is  \token{Protein} \token{input}}\\
   &&&&\texttt{\# \# Output: The descriptor value is \token{Descriptor} \token{output} }\\
   
   QED&2&MSE&MSE&\texttt{\# \# Input: The SMILES of the molecule is  \token{SMILES} \token{input}}\\
   &&&&\texttt{\# \# Output: The quantitative estimate of druglikeness is \token{QED} \token{output} }\\
   
  DC&2&MSE&MAE&\texttt{\# \# Input: Drug 1 is \token{input 0}. Its SMILES is \token{SMILES} \token{input 1}. }\\
  &&&&\texttt{\textcolor{white}{\# \# Input:} Drug 2 is \token{input 2}. Its SMILES is \token{SMILES} \token{input 3}}\\
   &&&&\texttt{\# \# Output: The drug combination sensitivity score is \token{DC} \token{output} }\\
   
  BA&4&MSE&Pearson&\texttt{\# \# Input: The protein sequence is \token{Protein} \token{input 0}. }\\
  &&&&\texttt{\textcolor{white}{\# \# Input:} The SMILES of the drug is \token{SMILES} \token{input 1}}\\
   &&&&\texttt{\# \# Output: The binding affinity  is \token{BA} \token{output} }\\
			\bottomrule
		\end{tabular}}
\label{appendix:table:task}
\end{table*}

\subsection{Experiment details: baselines}
\label{appendix:sec:baseline}

\subsubsection{Parameter efficient fine-tuning baselines}
\label{appendix:peftparam}
We use the \href{https://github.com/huggingface/peft}{Hugging Face PEFT library} to  implement the parameter-efficient fine-tuning baselines, including LoRA \citep{lora} and prompt tuning \citep{prompttuning}.
For fair comparison,  we use the same sets of training hyperparameters (e.g., number of epochs, batch size, learning rate, etc) as our method. Below, we detail method-specific configurations.
\begin{itemize}
    \item LoRA: rank=8, lora\_alpha=16, lora\_dropout=0.05
    \item Prompt tuning: for fair comparison, we set the num\_tokens field to the same as the total length of all tags inserted to the input (e.g., 20 for QED and 30 for BA); we use the PromptTuningInit.RANDOM initialization
    \item Linear probing: intialized with average embedding (same as our method)
\end{itemize}
We summarize the number of trainable parameters for different baselines in the following table (base LLM is LLaMA-7B). For our method, we assume using one domain tag and one function tag ($p=10$), which is the case for all single-domain tasks in our evaluation, such as QED, Descriptor, and DC. For prompt tuning, there are two virtual prompts for consistency with our method, and the length of each learnable prompt is also 10.
\begin{table}[h!]
\centering
\caption{Number of trainable parameters.}
\resizebox{0.5\textwidth}{!}{	\begin{tabular}{ccccc}
			\toprule
			    Base Model & Ours &	LoRA & Prompt Tuning & Linear Probing   \\
			\midrule
	 7B	&86016&1052672&86016&4096\\
			\bottomrule
		\end{tabular}}
\end{table}

\subsubsection{Nearest Neighbor Baseline}
We additionally implemented a nearest neighbor baseline for QED and Descriptor tasks. The basic idea is that, for each  data point in the test set, we want to find the most similar data point in the training set, and use the label of this most-similar data as our prediction. Since the data features are protein sequences and SMILES strings, we use the python SequenceMatcher function to select the nearest neighbor. However, this metric does not account for the intrinsic structure of the proteins and chemicals and thus performs poorly compared to the other learning-based approaches.

\subsubsection{Generation setup}
For both our method and the baseline, when evaluating with generation, we simply greedily decoded the most likely
sequence, as there are limited gains from using beam search with beam size $B = 4$.

\subsection{Tag-Mistral}
\label{appendix:mistral}
For the results reported in the main text, we mainly used LLaMA-7B as the LLM backbone to showcase the efficacy of our method. Here, we further investigate applying \Algo to another open-source model, Mistral-7B~\citep{jiang2023mistral}. The following results demonstrate the broad applicability of Tag-LLM beyond just the LLaMA models.

\begin{table*}[h!]
\label{table:mistral}
\renewcommand{\arraystretch}{1.1}
\centering
\begin{tabular}{lccccc}
		\toprule
   
 &  Descriptor  (MSE, $\downarrow$)  & QED  (MSE, $\downarrow$) &DC  (MAE, $\downarrow$)  &BA (Pearson $r$, $\uparrow$)\\  
\midrule
Tag-LLaMA	&0.005&\textbf{0.008}&	\textbf{12.21}	&	0.527\\
Tag-Mistral&\textbf{0.0048}&	0.011&	12.34		&0.472\\
Best Task-Specific Expert Model&	0.012	&0.04	&16.85&	\textbf{0.588}\\
\bottomrule
	\end{tabular}
\end{table*}

Across all four tasks, augmenting Mistral with input tags outperforms the PEFT baselines  and the  domain-specific LLM baselines in Table~\ref{table:proteindrug}, showing the effect of 1) injecting domain knowledge into the LLM via domain tags and 2) leveraging additional supervised data via function tags. Meanwhile, both Tag-Mistral and Tag-LLaMA perform comparably well. This demonstrates the compatibility of Tag-LLM with diverse pretrained backbones.


\end{document}